\documentclass[10pt,conference]{IEEEtran}
\IEEEoverridecommandlockouts
\usepackage{cite}
\usepackage{amsmath,amssymb,amsfonts}
\usepackage{algorithmic}
\usepackage{graphicx}
\usepackage{textcomp}

\usepackage[T1]{fontenc}
\usepackage{graphicx}
\usepackage{epsfig} 
\usepackage{mathptmx} 
\usepackage{times} 
\usepackage{amsmath} 
\usepackage{amssymb} 
\usepackage{multirow}
\usepackage{algorithm}
\usepackage{algorithmic}
\usepackage{multicol}
\usepackage{makecell}
\usepackage{enumerate}
\usepackage{hyperref}
\usepackage{booktabs} 
\usepackage[table]{xcolor}
\usepackage{listings}
\usepackage{balance}

\def\BibTeX{{\rm B\kern-.05em{\sc i\kern-.025em b}\kern-.08em
    T\kern-.1667em\lower.7ex\hbox{E}\kern-.125emX}}
\begin{document}

\title{Confusion-Aware In-Context-Learning for Vision-Language Models in Robotic Manipulation\\
}


\author{\IEEEauthorblockN{Yayun He}
\IEEEauthorblockA{\textit{Ping An Technology (Shenzhen) Co., Ltd.}\\
Shenzhen, China \\
heyayun0618@163.com}
\and
\IEEEauthorblockN{Zuheng Kang}
\IEEEauthorblockA{\textit{Ping An Technology (Shenzhen) Co., Ltd.}\\
Shenzhen, China \\
kangzuheng896@pingan.com.cn}
\and
\IEEEauthorblockN{Botao Zhao}
\IEEEauthorblockA{\textit{Ping An Technology (Shenzhen) Co., Ltd.}\\
Shenzhen, China \\
zhaobotao204@pingan.com.cn}
\and
\IEEEauthorblockN{Zhouyin Wu}
\IEEEauthorblockA{\textit{Shenzhen Bao'an Middle School}\\
Shenzhen, China \\
2789577758@qq.com}
\and
\IEEEauthorblockN{Junqing Peng}
\IEEEauthorblockA{\textit{Ping An Technology (Shenzhen) Co., Ltd.}\\
Shenzhen, China \\
pengjq@pingan.com.cn}
\and
\IEEEauthorblockN{Jianzong Wang$^*$}
\thanks{$^*$Corresponding author.}
\IEEEauthorblockA{\textit{Ping An Technology (Shenzhen) Co., Ltd.}\\
Shenzhen, China \\
jzwang@188.com}
}

\maketitle

\begin{abstract}
Vision-language models (VLMs) have significantly improved the generalization capabilities of robotic manipulation.
	However, VLM-based systems often suffer from a lack of robustness, leading to unpredictable errors, particularly in scenarios involving confusable objects.
	Our preliminary analysis reveals that these failures are mainly caused by shortcut learning problem inherently in VLMs, limiting their ability to accurately distinguish between confusable features.
	To this end, we propose Confusion-Aware In-Context Learning (CAICL), a method that enhances VLM performance in confusable scenarios for robotic manipulation.
	The approach begins with confusion localization and analysis, identifying potential sources of confusion.
	This information is then used as a prompt for the VLM to focus on features most likely to cause misidentification.
	Extensive experiments on the VIMA-Bench show that CAICL effectively addresses the shortcut learning issue, achieving a 85.5\% success rate and showing good stability across tasks with different degrees of generalization.
\end{abstract}


\section{Introduction}
Vision plays a crucial role in human perception, with over 90\% of information acquired through sight \cite{liu2024integration}.
Integrating computer vision (CV) into robotics enhances their ability to perceive, interpret, and interact intelligently with their environment and humans.
In scenarios like manufacturing and healthcare, CV-equipped robots have proven to be highly effective in improving efficiency and precision.
For instance, in manufacturing, they enhance productivity through precise object handling \cite{golnabi2007design}, while in healthcare, they assist in critical tasks like surgery, reducing the risk of errors \cite{yang2020homecare}.
Various visual methods have been developed to aid robotic tasks, including the YOLO series \cite{bochkovskiy2020yolov4,redmon2016you} and RCNN \cite{girshick2014rich,ren2016faster} for image recognition, and U-Net \cite{ronneberger2015u} for image segmentation.
However, these models often lack generalizability and require fine-tuning for adaptation to new environments.
More recent models, such as the segment anything model (SAM) \cite{kirillov2023segment} and CLIP \cite{radford2021learning,hafner2021clip}, show improvements in generalizability and robustness across a range of tasks.
Despite these advances, challenges persist in handling more complex environments and intricate visual tasks.
Additionally, many models lack iterative reasoning capabilities, limiting their ability to refine and adjust outputs through step-by-step processes.

Recent advances in vision-language models (VLMs) \cite{zhu2023minigpt,liu2023llava,liu2024llavanext} have significantly improved robotic capabilities in tasks such as visual question answering (VQA) and image captioning \cite{lin2023medical,shao2023prompting}.
These models leverage multimodal pretraining to align visual and textual data, enabling robots to interpret complex scenes and generate contextually relevant responses.
For instance, in VQA tasks, VLMs allow robots to analyze visual inputs and provide accurate answers to user queries, while in image captioning, they generate descriptive summaries of visual content.
By integrating visual and linguistic information, VLMs improve a robot's ability to reason comprehensively, especially in complex command navigation \cite{zhou2024navgpt}.
However, VLM-based robotic manipulation achieves strong generalization at the expense of robustness, which often leads to unpredictable failures.
Robustness is a critical factor in domains such as healthcare and industrial applications, where insufficient operational stability can result in irreversible consequences.
Although various prompt strategies have been developed to optimize VLM performance \cite{yao2023react,wei2022chain,wangself}, challenges remain in ensuring reliable and consistent performance across diverse environments and tasks.

\begin{figure*}[t]
	\centering
	\includegraphics[width=0.90\textwidth]{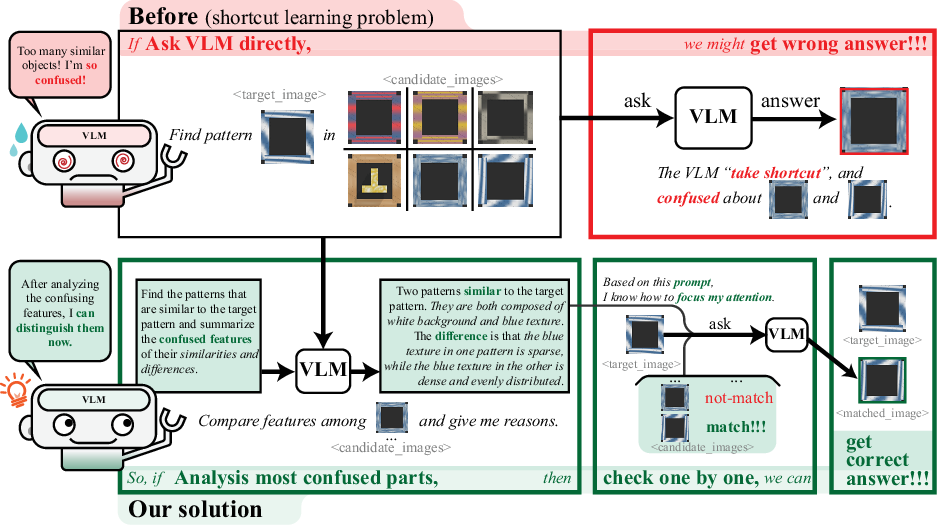}
	\caption{Illustration of Shortcut Learning in Vision-Language Models (VLM) and the Confusion-Aware In-Context Learning (CAICL) Approach.
		The upper part shows the VLM making incorrect decisions due to confusion between similar objects, a result of shortcut learning.
		The proposed CAICL method, however, guides the VLM to scan the entire scene, identify confusable features, and improve image recognition accuracy.}
	\label{fig:motivation}
	\vspace{-0.5em}
\end{figure*}

In this paper, we designed experiments to identify the primary cause of failures in VLM-based robotic manipulation systems.
Our findings reveal that most failures occur during the image recognition phase of environmental perception. Further analysis indicates that these failures stem from the shortcut learning problem \cite{sun2024exploring,bleeker2024demonstrating}, where VLMs tend to make quick, superficial judgments rather than deeply analyzing the scene.
As illustrated in Figure \ref{fig:motivation}, this tendency to \textbf{``take shortcuts''} prevents VLMs from accurately distinguishing between visually similar objects, leading to frequent misidentification.
This behavior suggests that VLMs lack a mechanism to effectively direct their attention toward the most confusable aspects of a complicated scene.
To overcome this limitation, we propose a recognition approach that dynamically identifies confusable features, analyzes them, and refocuses the VLM's attention to improve accuracy and reduce confusion.

To address the above issues, we have made the following contributions:
\vspace{1em}
\begin{itemize}
	\item We identified the shortcut learning problem that exists in VLM-based robotic manipulation systems, particularly in image recognition, which impairs accurate environmental understanding.
	\item We propose a confusion-aware in-context learning (CAICL) method to address shortcut learning issues.
	      This prompt-based framework shifts the model's attention to the most confusable elements in the scene, thereby improving recognition accuracy.
	\item Extensive experiments on 12 tasks from the VIMA-Bench demonstrate that our method mitigates shortcut learning, achieving state-of-the-art performance.
\end{itemize}

\vspace{1em}
\section{Related Works}
\subsection{VLMs for Robotic Manipulation}

Robotic manipulation involves performing specific tasks through interaction with the environment.
However, the complexity of these tasks requires large amounts of training data to ensure robust generalization, which demands significant time and computational resources \cite{fang2023rh20tcomprehensiveroboticdataset,shridhar2022perceiveractormultitasktransformerrobotic}.
As a result, there has been increasing interest in using large language models (LLMs) and VLMs to enhance the generalization capabilities of robotic manipulation.
For instance, SayCAN~\cite{ahn2022icanisay} combines LLMs with pretrained robotic skills and value functions to map natural language instructions to robotic actions.
Palm-E~\cite{driess2023palme} integrates both visual and language modules to enable more effective task understanding and execution, demonstrating the potential of VLMs to bridge the gap between language and physical actions.
Similarly, RT-2~\cite{zitkovich2023rt2} utilizes vision-language-action (VLA) models to transfer web knowledge for improved task generalization and semantic reasoning.
CaP~\cite{liang2023code} and Instruct2Act~\cite{huang2023instruct2act} leverage VLMs to generate executable action sequences for robotic manipulation, showing promise in task adaptability.

Despite these advancements, VLM-based systems often face challenges related to robustness, particularly in scenarios involving confusable objects.
The complexity and variability of real-world environments can lead to errors, as VLMs may struggle to distinguish between visually similar features \cite{radford2021learning,kirillov2023segment}.
This issue is further named by ``shortcut learning'' \cite{wei2022chain}.

\begin{figure*}[t]
	\centering
	\includegraphics[width=0.9\textwidth]{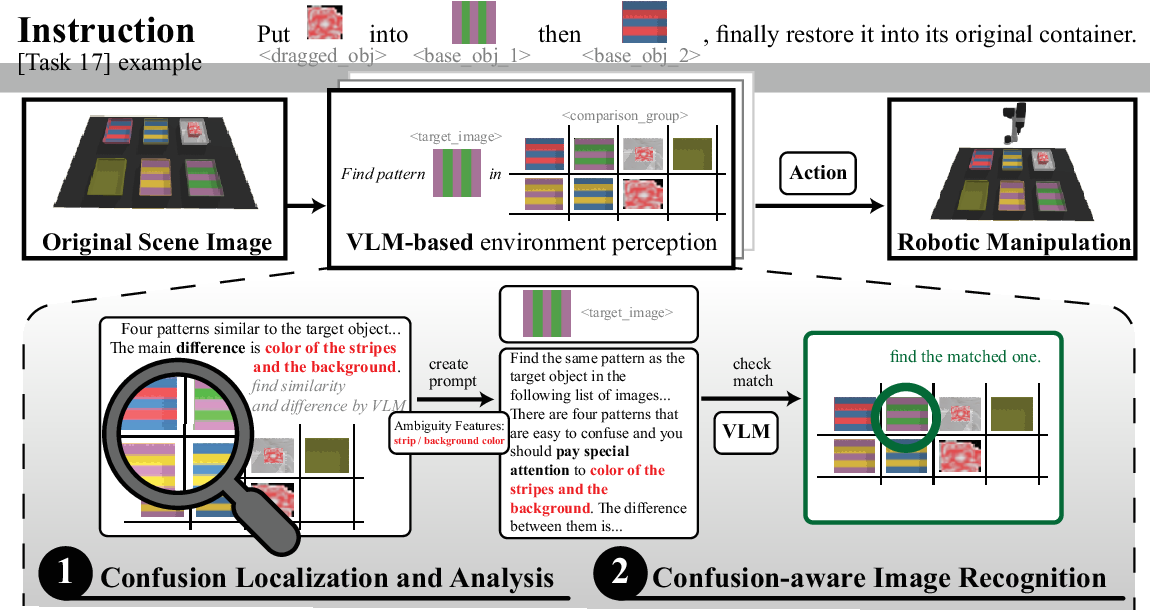 }
	\caption{The overall process of the confusion-aware in-context-learning (CAICL).}
	\label{fig:main}
	\vspace{-0.5em}
\end{figure*}

\vspace{0.5em}
\subsection{Shortcut Learning Problem}

Shortcut learning refers to the tendency of models to rely on superficial correlations or simple features rather than capturing deeper, underlying patterns, ultimately limiting their generalization and robustness \cite{geirhos2020shortcut}.
This phenomenon has been widely studied in the context of LLMs, where models often exploit spurious correlations such as high-frequency labels \cite{zhao2021calibrate}, prompt order \cite{lu2022fantastically}, domain-specific knowledge \cite{razeghi2022impact}, and lexical features \cite{tang2023large}.
Studies have shown that these biases stem from the co-occurrence of words and labels in pretraining data, leading to unreliable performance on more challenging tasks \cite{kang2023impact,mccoy2023embers}.
For example, LLMs may prefer high-frequency labels regardless of context \cite{zhao2021calibrate} or exhibit sensitivity to option order in multiple-choice questions \cite{pezeshkpour2024large}.
Additionally, LLMs can exhibit biases related to the format and structure of the input text \cite{sclar2024quantifying}.

To fully address this problem, we conducted various mitigation strategies have been proposed to address shortcut learning in LLMs.
Some studies focus on resampling and retraining to balance data distributions \cite{kang2023impact}, while leverage model pruning to remove biased neurons \cite{yang2024mitigating}.
Calibration techniques, such as contextual calibration \cite{zhao2021calibrate} and batch calibration \cite{zhou2024batch}, adjust prediction distributions to counteract biases.
Other methods like synthetic oversampling to mitigate data imbalance \cite{nakada2024synthetic} and prompt-based strategies to gain more robust reasoning \cite{zhou2023navigating,levy2023guiding}.

However, although shortcut learning has been extensively explored in LLMs, its implications for VLMs remain underexamined.
Unlike LLMs, VLMs integrate both visual and textual information, which introduces new complexities in dealing with shortcut learning.
In VLMs, shortcuts can emerge from either modality or their interaction, posing unique challenges.
For instance, a VLM might over-rely on superficial visual features or specific text patterns frequently associated with certain labels in the training data, leading to poor generalization in new scenarios \cite{du2024shortcut}.
Therefore, our research aims to investigate the distinct characteristics and causes of shortcut learning in VLMs and propose strategies to mitigate these issues, improving their robustness and generalization in visual-language tasks.

\begin{figure*}[t]
	\centering
	\includegraphics[width=0.90\textwidth]{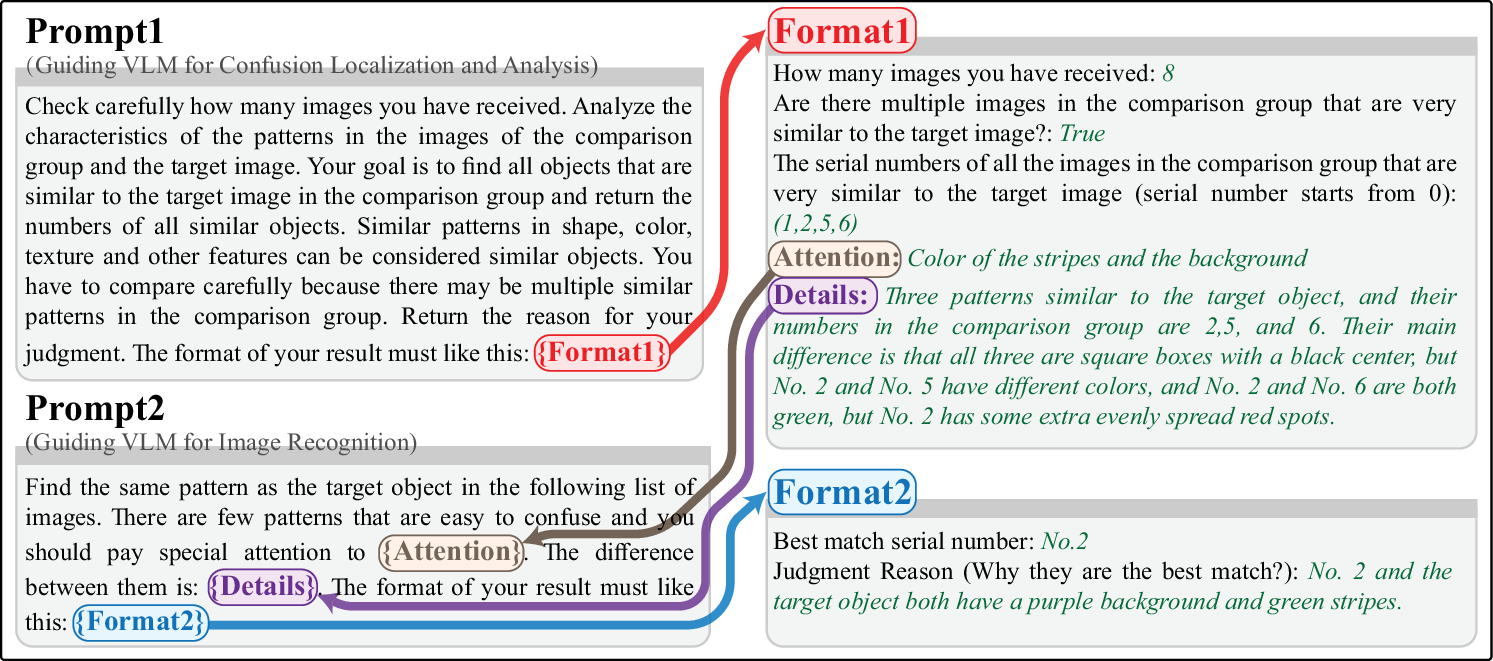}
	\caption{Prompts used for CAICL in confusion location and image recognition.}
	\label{fig:prompt}
	\vspace{-0.5em}
\end{figure*}

\section{Methodology}

\subsection{Shortcut Learning Problem in VLMs For Robotic Manipulation}

Robotic manipulation is a challenging engineering problem, where errors at any stage can result in the failure of the overall task.
To investigate the impact of VLMs in this context, we developed a manipulation pipeline based on Instruct2Act \cite{huang2023instruct2act}, which generates executable code and uses CLIP for image recognition to produce action sequences from environmental inputs and instructions.
In our implementation, we replaced the CLIP-based image recognition with VLM-driven question-answering pairs.

\begin{figure}[t]
	\centering
	\includegraphics[width=0.4\textwidth]{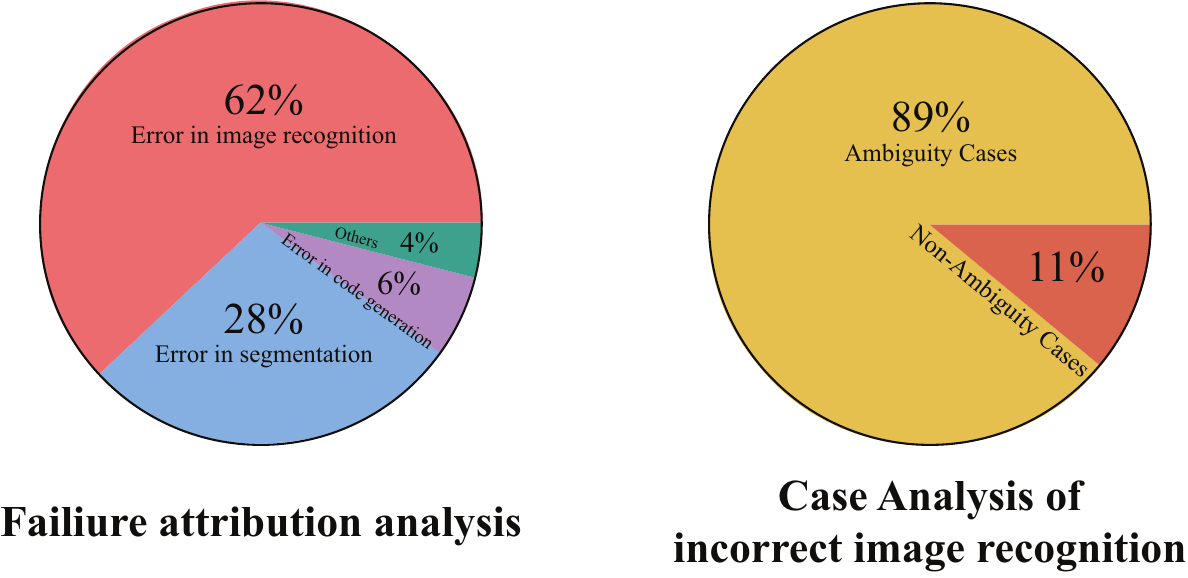}
	\caption{The left subplot shows the proportion of the failure cases on the VIMA-Bench Task 17.
		The right plot illustrates the proportion of the confusion cases and confusion cases for the incorrect image recognition cases.}
	\label{fig2-analysis}
\end{figure}

We evaluated the performance of this modified pipeline on the tasks of VIMA-Bench, at the L1, L2, and L3 generalization levels, and conducted a manual attribution analysis of failed tasks.
For example, in task 17 (Figure \ref{fig2-analysis}), the main failure sources were errors in image segmentation, image recognition, and code generation.
Notably, incorrect image recognition accounted for about 60\% of the failures, making it the leading cause.
Further analysis revealed that many of these errors were due to confusion, where the model selected the second-best match instead of the correct one.
This indicates that VLMs are prone to a ``slack'' phenomenon, similar to ``shortcut learning'' \cite{sun2024exploring,bleeker2024demonstrating}, where the model makes quick, suboptimal decisions based on partial information, similar to a human choosing an answer in a multiple-choice question without thoroughly evaluating all options.
This tendency presents a significant challenge for VLMs in robotic manipulation tasks.

\subsection{Confusion-Aware In-Context Learning}

To address the issue of shortcut learning in VLMs, we propose confusion-aware in-context learning (CAICL) for improving performance in tasks prone to being confused.
As shown in Figure \ref{fig:main}, CAICL consists of two main phases:
(1) confusion localization and analysis,
(2) confusion-aware image recognition.
In the first phase, the model identifies features contributing to confusion among comparison images, while in the second, it uses these features to refine its image recognition process.

\subsubsection{Confusion Localization and Analysis}
\label{sec:local}

This phase serves as a pre-recognition step where the model scans the scene to identify objects that may cause visual confusion with the target.
These visually similar objects can hinder the model's ability to accurately recognize the target.
The process involves two key steps:

\begin{enumerate}
	\item \textbf{Scene scanning and identification of confusable candidates:}
	      The VLM begins by thoroughly scanning the entire scene to identify all objects present.
	      During this scan, the model flags objects that appear visually similar to the target.
	      The focus is primarily on the elemental composition and color distribution of the objects, while the orientation or pose of the objects is not considered at this stage.
	\item \textbf{Confusion feature extraction:}
	      After identifying potential confusion candidates, the VLM extracts distinctive features that differentiate them, such as subtle texture variations or minor geometric differences.
	      This extracted information serves as crucial guidance for the subsequent recognition phase, helping the model to achieve more accurate environmental perception and minimize visual confusion.
\end{enumerate}

Consider the example in Figure \ref{fig:main} and Figure \ref{fig:prompt}, where we use \texttt{prompt1} to instruct the VLM to compare the most confusable features in \texttt{format1}.
This process begins by asking the VLM for specific details to help it fully understand the environment, such as the number of images present, the most confusable images (with their serial numbers), and the reasons for the confusion.
Finally, we request special \texttt{attention} and \texttt{details} about the confusable features that the VLM has learned.
This approach encourages the VLM to think more critically, compare detailed information across the environment, and identify the most confusable features in all images and backgrounds.

\subsubsection{Confusion-Aware Image Recognition}

In the second phase, we introduce a confusion-aware reasoning approach to enhance image recognition.
This method builds upon the identification of confusable features, prompting the model to focus specifically on resolving these confusions.

As demonstrated in Figure \ref{fig:main} and Figure \ref{fig:prompt}, rather than simply selecting the most similar image, the model is required to provide a detailed reasoning process.
This reasoning is guided by the \textbf{confusion localization and analysis} step (described in \ref{sec:local}), which identifies the key features contributing to the confusion, along with their corresponding \texttt{attention} and \texttt{details} descriptions.
The output of this analysis is then integrated into \texttt{prompt2}, which asks the model to specify the serial number of the best matching image and justify its choice.
This forces the model to reconsider and focus on the most confusable features, guiding it toward more accurate judgments.
By explicitly guiding the model to examine the most confusable features and generate structured, comparative outputs, the CAICL approach helps reduce the risks of shortcut learning, which can lead to errors in complex scenes.

\begin{table*}[t]
	\centering
    \scriptsize
	\caption{The comparison of the success rate on VIMA-Bench across L1, L2, and L3 generalization with 17 tasks.}
	\setlength{\tabcolsep}{6.8pt}
	\renewcommand{\arraystretch}{0.96}
	\begin{tabular}{lcccccccccccccccccc}
		\toprule
		\multirow{2}{*}{Method} & \multicolumn{17}{c}{Tasks} & \multirow{2}{*}{Avg.}                                                                                                                    \\
		\cmidrule(lr){2-18}
		                        & 01                         & 02                    & 03    & 04    & 05   & 06    & 07    & 08 & 09   & 10 & 11   & 12 & 13 & 14 & 15   & 16   & 17                   \\
		\midrule
		\rowcolor{yellow!20}
		\multicolumn{19}{c}{\textbf{L1 Level}}                                                                                                                                                          \\
		\midrule
		\textbf{Gato}           & 79.0                       & 68.0                  & 91.5  & 57.0  & 44.5 & 54.0  & 74.0  & -  & 18.0 & -  & 61.0 & -  & -  & -  & 83.5 & 33.5 & 2.5  & 58.1          \\
		\textbf{Flamingo}       & 56.0                       & 58.5                  & 63.0  & 48.5  & 38.0 & 48.5  & 62.5  & -  & 3.5  & -  & 66.5 & -  & -  & -  & 40.0 & 43.5 & 2.5  & 47.5          \\
		\textbf{GPT}            & 62.0                       & 57.5                  & 41.0  & 55.5  & 45.5 & 47.5  & 54.5  & -  & 8.5  & -  & 77.0 & -  & -  & -  & 41.0 & 38.0 & 0.5  & 46.9          \\
		\textbf{VIMA}           & 100.0                      & 100.0                 & 99.5  & 100.0 & 56.5 & 100.0 & 100.0 & -  & 18.0 & -  & 77.0 & -  & -  & -  & 97.0 & 76.5 & 43.0 & 81.6          \\
		\textbf{RT-2}           & 100.0                      & 98.0                  & 97.0  & 58.0  & 30.0 & 98.0  & 97.0  & -  & 14.0 & -  & 84.0 & -  & -  & -  & 93.0 & 52.0 & 47.0 & 72.8          \\
		\textbf{CaP}            & 90.0                       & 80.8                  & 96.0  & 65.3  & 61.3 & 80.0  & 84.7  & -  & 38.0 & -  & 68.0 & -  & -  & -  & 67.3 & 58.0 & 76.0 & 71.2          \\
		\textbf{VisualProg}     & 92.0                       & 27.0                  & 9.0   & 29.0  & 90.0 & 38.0  & 87.0  & -  & 21.3 & -  & 65.3 & -  & -  & -  & 92.0 & 36.7 & 28.7 & 49.7          \\
		\textbf{CLIP(I2A)}      & 91.3                       & 81.4                  & 98.2  & 78.5  & 72.0 & 82.0  & 88.0  & -  & 42.0 & -  & 72.0 & -  & -  & -  & 78.0 & 64.0 & 85.2 & 77.0          \\
		\textbf{Vanilla VLM}    & 87.9                       & 89.6                  & 97.8  & 75.3  & 73.2 & 82.8  & 92.2  & -  & 46.0 & -  & 81.1 & -  & -  & -  & 84.6 & 78.3 & 84.1 & 81.2          \\
		\rowcolor{gray!20}
		\textbf{CAICL(Ours)}    & 93.1                       & 92.5                  & 98.8  & 86.2  & 85.5 & 84.9  & 91.6  & -  & 50.1 & -  & 83.3 & -  & -  & -  & 85.2 & 82.4 & 92.2 & \textbf{85.5} \\
		\midrule
		\rowcolor{yellow!20}
		\multicolumn{19}{c}{\textbf{L2 Level}}                                                                                                                                                          \\
		\midrule
		\textbf{Gato}           & 56.5                       & 53.5                  & 88.0  & 55.5  & 43.5 & 55.5  & 53.0  & -  & 14.0 & -  & 63.0 & -  & -  & -  & 81.5 & 33.0 & 4.0  & 53.2          \\
		\textbf{Flamingo}       & 51.0                       & 52.5                  & 61.5  & 49.5  & 38.5 & 47.5  & 55.5  & -  & 5.5  & -  & 70.5 & -  & -  & -  & 42.0 & 39.0 & 3.0  & 46.0          \\
		\textbf{GPT}            & 52.0                       & 52.0                  & 49.5  & 54.5  & 45.5 & 52.5  & 51.0  & -  & 11.0 & -  & 76.5 & -  & -  & -  & 43.0 & 38.0 & 0.5  & 46.9          \\
		\textbf{VIMA}           & 100.0                      & 100.0                 & 99.5  & 100.0 & 54.5 & 100.0 & 100.0 & -  & 17.5 & -  & 77.0 & -  & -  & -  & 98.5 & 75.0 & 45.0 & 81.5          \\
		\textbf{RT-2}           & 100.0                      & 96.0                  & 97.0  & 56.0  & 27.0 & 95.0  & 97.0  & -  & 10.0 & -  & 84.0 & -  & -  & -  & 92.0 & 43.0 & 34.0 & 70.3          \\
		\textbf{CaP}            & 90.0                       & 79.3                  & 96.0  & 64.7  & 60.0 & 74.6  & 85.3  & -  & 37.3 & -  & 66.7 & -  & -  & -  & 66.0 & 52.7 & 74.7 & 70.0          \\
		\textbf{VisualProg}     & 84.0                       & 26.0                  & 11.3  & 38.7  & 87.3 & 30.0  & 80.7  & -  & 20.0 & -  & 71.3 & -  & -  & -  & 94.7 & 24.0 & 30.0 & 47.7          \\
		\textbf{CLIP(I2A)}      & 91.5                       & 80.8                  & 97.8  & 74.9  & 69.5 & 81.0  & 86.0  & -  & 44.0 & -  & 70.5 & -  & -  & -  & 80.0 & 66.0 & 84.0 & 76.2          \\
		\textbf{Vanilla VLM}    & 88.1                       & 89.4                  & 97.6  & 74.3  & 71.8 & 80.3  & 91.2  & -  & 47.1 & -  & 80.3 & -  & -  & -  & 85.7 & 77.7 & 83.1 & 80.5          \\
		\rowcolor{gray!20}
		\textbf{CAICL(Ours)}    & 94.6                       & 92.2                  & 99.3  & 85.7  & 83.5 & 83.3  & 90.6  & -  & 51.0 & -  & 82.2 & -  & -  & -  & 85.2 & 81.1 & 90.7 & \textbf{85.0} \\
		\midrule
		\rowcolor{yellow!20}
		\multicolumn{19}{c}{\textbf{L3 Level}}                                                                                                                                                          \\
		\midrule
		\textbf{Gato}           & 51.0                       & 58.0                  & 84.5  & 56.5  & 35.5 & 53.5  & 49.0  & -  & 15.0 & -  & 65.0 & -  & -  & -  & 52.0 & 33.0 & 0.0  & 43.5          \\
		\textbf{Flamingo}       & 49.0                       & 50.0                  & 66.5  & 47.0  & 35.0 & 47.5  & 50.0  & -  & 4.0  & -  & 66.0 & -  & -  & -  & 30.5 & 43.5 & 0.5  & 40.8          \\
		\textbf{GPT}            & 52.0                       & 51.0                  & 55.0  & 49.5  & 40.0 & 46.0  & 50.5  & -  & 5.0  & -  & 82.0 & -  & -  & -  & 37.0 & 38.0 & 1.5  & 42.3          \\
		\textbf{VIMA}           & 99.0                       & 100.0                 & 100.0 & 97.0  & 58.0 & 100.0 & 99.0  & -  & 17.5 & -  & 90.5 & -  & -  & -  & 97.5 & 46.0 & 43.5 & 79.0          \\
		\textbf{RT-2}           & 96.0                       & 94.0                  & 96.0  & 52.0  & 31.0 & 95.0  & 93.0  & -  & 11.0 & -  & 97.0 & -  & -  & -  & 93.0 & 40.0 & 3.0  & 66.8          \\
		\textbf{CaP}            & 90.0                       & 79.3                  & 95.3  & 63.3  & 60.0 & 74.0  & 84.7  & -  & 37.3 & -  & 66.0 & -  & -  & -  & 64.3 & 51.3 & 72.0 & 69.8          \\
		\textbf{VisualProg}     & 83.3                       & 25.3                  & 13.3  & 27.3  & 62.0 & 32.0  & 80.7  & -  & 17.3 & -  & 70.0 & -  & -  & -  & 73.3 & 24.0 & 39.7 & 45.7          \\
		\textbf{CLIP(I2A)}      & 91.8                       & 80.2                  & 97.4  & 81.8  & 65.8 & 79.0  & 89.0  & -  & 38.0 & -  & 71.0 & -  & -  & -  & 78.0 & 62.0 & 82.0 & 76.3          \\
		\textbf{Vanilla VLM}    & 88.2                       & 89.6                  & 97.8  & 75.8  & 68.7 & 79.8  & 92.8  & -  & 45.2 & -  & 80.5 & -  & -  & -  & 84.4 & 78.6 & 81.7 & 80.2          \\
		\rowcolor{gray!20}
		\textbf{CAICL(Ours)}    & 93.7                       & 92.8                  & 98.9  & 86.7  & 81.5 & 83.8  & 93.2  & -  & 48.6 & -  & 83.2 & -  & -  & -  & 84.9 & 81.6 & 90.4 & \textbf{84.9} \\
		\bottomrule
	\end{tabular}
	\label{tab:overall}
	\vspace{-1em}
\end{table*}

\begin{figure*}[htb]
	\centering
	\includegraphics[width=0.99\textwidth]{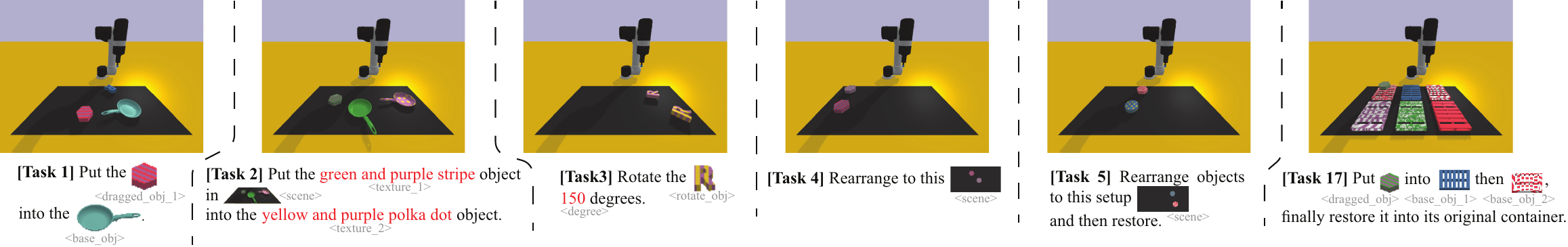}
	\caption{Some selected tasks of VIMA-Bench.}
	\label{fig:tasks}
\end{figure*}

\section{Experiments}

\subsection{Experimental Setup}

To evaluate our proposed approach, we used the VIMA-Bench dataset \cite{jiang2022vima}, which contains 17 tasks covering a wide range of operations, including basic object manipulations (e.g., Task 1, Task 3), complex multi-step operations (e.g., Task 5, Task 17), scene understanding (Task 2) and visual goal matching (Task 04).
These tasks encompass common challenges in robotic manipulation, including object recognition, goal matching, multi-step planning, and restoration.
To better understand these challenges, we selected seven representative tasks, as shown in Figure \ref{fig:tasks}, to evaluate their specific characteristics.
Also, the VIMA-Bench dataset includes a four-tier generalization evaluation protocol: L1 (placement generalization), L2 (combinatorial generalization), L3 (novel object generalization), and L4 (novel task generalization).
Since L4 is not applicable to many tasks, our experiments focus on generalization tasks L1, L2 and L3.
Notably, due to the similarity between some tasks and the absence of tasks 08, 10, and 12-14 in generalization levels L1 to L3, we selected the remaining 12 tasks for our evaluation, consistent with \cite{jiang2022vima}.
Our implementation follows the Instruct2Act framework for the code generation and action planning part, while the environment perception part is built using the GPT-4o interface, integrated with the large version of SAM.
We evaluate the performance of our approach using the success rate (SR).

\subsection{Baselines}

To comprehensively evaluate our approach, we compared it with several baseline methods in two main categories.

\subsubsection{Learning-based Methods}

(a) VIMA \cite{jiang2022vima} integrates instruction and observation data through multimodal prompts and cross-attention blocks;
(b) Gato \cite{reedgeneralist} uses a decoder-only architecture that processes observation and action subsequences with prompting;
(c) Flamingo \cite{alayrac2022flamingo} encodes a variable number of prompt images into a fixed number of tokens, similar to the Perceiver architecture, and includes robot action heads;
(d) GPT \cite{chen2021decision} is a decoder-only model that autoregressively generates actions from tokenized multimodal prompts, encoding images into state tokens using a ViT encoder, without cross-attention;
(e) RT-2 \cite{zitkovich2023rt2} trains a large robotic model through stacked multi-transformer decoder blocks for autoregressive generation.

\subsubsection{Prompt-based Methods}

(a) VisProg generates modular, python-like programs that are executed to produce both a solution and an interpretable rationale;
(b) CaP \cite{liang2023code} generates code as policies to control the robot based on textual instructions;
(c) Replace the image recognition module with CLIP, similar to Instruct2Act;
(d) The Vanilla VLM approach, which directly poses questions to the vision-language model (VLM) without CAICL, as shown in the upper part of Figure \ref{fig:motivation}.

\section{Results}

\subsection{Main Performance}

As shown in Table \ref{tab:overall}, we evaluated the performance of various methods on VIMA-Bench across 12 tasks at three generalization levels: L1, L2, and L3. Our proposed method, CAICL, outperforms all baseline models with average success rates of 85.5\% (L1), 85.0\% (L2), and 84.9\% (L3), demonstrating its superior performance overall.
Compared to VIMA, the best-performing traditional learning-based method, CAICL shows significant improvements in success rates across all generalization levels.
Additionally, when comparing vanilla VLM with CLIP (I2A), we observe a slight decrease in accuracy for vanilla VLM on some tasks.
However, its overall average accuracy improves from 77.0\% (CLIP) to 81.2\% (Vanilla VLM) at L1, indicating that VLMs effectively enhance visual perception for robotic manipulation tasks.
This improvement highlights the value of large-scale vision-language pretraining for task comprehension and execution.

Our proposed method further builds on the vanilla VLM by incorporating Confusion-Aware In-Context Learning (CAICL), resulting in higher average accuracy and superior performance across all 12 tasks.
This indicates that CAICL effectively mitigates the shortcut learning issue, which typically weakens VLMs in object discrimination tasks.
Notably, CAICL maintains stable performance as generalization levels increase, demonstrating its strong generalization capabilities and robustness in more complicated scenarios.

In summary, our experimental results validate the effectiveness of CAICL in improving the performance of VLMs for robotic manipulation tasks.
The method not only outperforms existing baselines but also addresses key challenges such as shortcut learning and generalization, making it a promising approach for future research in this domain.

\begin{table}[t]
	\centering
	\caption{The average success rate on the VIMA-Bench across L1, L2, and L3 generalization when using various VLM.}
	\setlength{\tabcolsep}{5.5pt}
	\renewcommand{\arraystretch}{1.0}
	\begin{tabular}{lcccccc}
		\toprule
		\multicolumn{1}{l}{\multirow{1}{*}{Methods}} & Task1         & Task2         & Task3         & Task4         & Task5         & Task17        \\
		\midrule
		LLaVA-v1.5                                   & 80.2          & 79.4          & 83.2          & 74.5          & 70.7          & 69.4          \\
		Qwen2-VL                                     & 85.0          & 84.4          & 87.8          & 79.7          & 76.8          & 78.4          \\
		Gemini 1.5 Pro                               & 93.5          & \textbf{93.2} & \textbf{99.0} & \textbf{88.4} & 83.4          & 91.2          \\
		GPT-4o                                       & \textbf{93.8} & 92.5          & \textbf{99.0} & 86.2          & \textbf{83.5} & \textbf{91.7} \\
		\bottomrule
	\end{tabular}
	\label{tab:ablation}
\end{table}

\vspace{-0.5em}
\subsection{Ablation Studies}

Table \ref{tab:ablation} presents the results of our ablation study, evaluating the performance of different VLMs in a robotic manipulation system.
Due to computational constraints, we selected the most representative tasks (Task 1, 2, 3, 4, 5, and 17) for evaluation.
Smaller models, such as LLaVA-v1.5 (7B), achieved moderate success rates (80.2\% on Task 1 and 69.4\% on Task 17).
In contrast, larger models like Qwen2-VL (70B) showed significant improvements, with success rates of 85.0\% on Task 1 and 78.4\% on Task 17.
The recent state-of-the-art models, Gemini 1.5 Pro and GPT-4o, delivered the highest performance.
Gemini excelled on Task 2 (93.2\%), Task 3 (99.0\%), and Task 4 (88.4\%), while GPT-4o led on Task 1 (93.8\%), Task 5 (83.5\%), and Task 17 (91.7\%).

These results highlight the significant impact of model scale and advanced capabilities, such as improved reasoning and contextual understanding, on robotic manipulation performance.
While smaller models offer a baseline, larger and more sophisticated models greatly enhance task accuracy and robustness, demonstrating their potential for real-world applications.

\vspace{-0.5em}
\section{Conclusion}
In conclusion, this paper identifies the primary challenge in VLM-based robotic manipulation systems as image recognition failures during environmental understanding, particularly in scenarios involving visually confusable objects.
Our analysis attributes these failures to the shortcut learning tendencies of VLMs, especially in distinguishing between similar features.
To address this, we proposed confusion-aware in-context learning (CAICL), which encourages deeper reasoning by prompting the model to focus on confusable features and avoid superficial judgments.
Our approach leverages VLMs to generate descriptive texts that emphasize confusable features, prompting the model to focus on subsequent recognition tasks, particularly in regions prone to misidentification.
Validated on tasks of VIMA-Bench, our approach demonstrates its effectiveness in mitigating the shortcut learning problem and enhancing the robustness of VLM-based robotic manipulation systems.

\section{Acknowledgement}
This work was supported by Shenzhen-Hong Kong Joint Funding Project (Category A) under Grant No. SGDX20240115103359001.


\balance
\bibliographystyle{unsrt}
\bibliography{refs}

\end{document}